\documentclass[conference]{IEEEtran}
\IEEEoverridecommandlockouts
\usepackage{cite}
\usepackage[utf8]{inputenc}
\usepackage{graphicx}
\usepackage{listings}
\usepackage[dvipsnames]{xcolor}
\usepackage{float}
\usepackage{booktabs}
\usepackage{algorithmic, algorithm, }
\usepackage{threeparttable}
\usepackage{amsmath, amsfonts, amsthm, amssymb, mathtools, xparse, stackengine}
\usepackage{footnote}
\makesavenoteenv{threeparttable}
\usepackage{tablefootnote}
\stackMath

\definecolor{custom_blue}{HTML}{1F77B4}
\definecolor{custom_orange}{HTML}{FF7F0E}

\usepackage{hyperref}

\usepackage[utf8]{inputenc}
\usepackage[margin=1in]{geometry}

\usepackage{subfigure}

\def\BibTeX{{\rm B\kern-.05em{\sc i\kern-.025em b}\kern-.08em
    T\kern-.1667em\lower.7ex\hbox{E}\kern-.125emX}}
    
\begin{document}

\author{
    \IEEEauthorblockN{Baturay Saglam, Enes Duran, Dogan C. Cicek, Furkan B. Mutlu \\ and Suleyman S. Kozat, \IEEEmembership{Senior~Member,~IEEE}} \\
    \IEEEauthorblockA{Department of Electrical and Electronics Engineering, Bilkent University, Ankara, Turkey}
    \IEEEauthorblockA{\tt\footnotesize\{baturay, enesd, cicek, kozat\}@ee.bilkent.edu.tr} \tt\footnotesize\{burak.mutlu\}@bilkent.edu.tr}
    
\title{Estimation Error Correction in Deep Reinforcement Learning for Deterministic Actor-Critic Methods}

\maketitle

\begin{abstract}
In value-based deep reinforcement learning methods, approximation of value functions induces overestimation bias and leads to suboptimal policies. We show that in deep actor-critic methods that aim to overcome the overestimation bias, if the reinforcement signals received by the agent have a high variance, a significant underestimation bias arises. To minimize the underestimation, we introduce a parameter-free, novel deep Q-learning variant. Our Q-value update rule combines the notions behind Clipped Double Q-learning and Maxmin Q-learning by computing the critic objective through the nested combination of maximum and minimum operators to bound the approximate value estimates. We evaluate our modification on the suite of several OpenAI Gym continuous control tasks, improving the state-of-the-art in every environment tested.
\end{abstract}


\section{Introduction}
\subsection{Preliminaries}
In recent years, utilization of deep approaches to approximate the policies of reinforcement learning (RL) agents achieved numerous successes in wide range of applications such as playing Atari games \cite{dqn}, autonomous driving \cite{autonomous_driving}, path planning \cite{rrt_rl}, playing board games like chess, shogi \cite{chess}, go \cite{go} and even beating human players on StarCraft \cite{starcraft} \cite{openai_five}. Nevertheless, there are several issues regarding the function approximation in the deep reinforcement learning setting \cite{td3}. One of the problems resulting from the function approximation is the systematic estimation bias that prevents the learning agent from reaching the maximum performance and deep methods to be applied to the real-world problems \cite{sutton88}, \cite{off_policy}. The estimation bias on value estimates in value-based RL algorithms has been studied for discrete action spaces \cite{ddqn, insights_rl, precup_2001, max_min, averaged_dqn}. Furthermore, similar work in the continuous control domain for actor-critic methods is done for a subcategory of estimation bias, that is, overestimation bias \cite{td3}. This paper shows that in deterministic actor-critic methods that aim to overcome the accumulated overestimation bias and high variance, there exists an underestimation bias on the value estimates \cite{wd3}, \cite{tadd}. Our work addresses this issue from a probabilistic point of view and improves the current state-of-the-art performance on several continuous control RL tasks.

The estimation bias on the action value estimates in value-based deep reinforcement learning is usually studied in two categories: underestimation and overestimation. Overestimation bias, resulting from the maximization of noisy estimates in the standard Q-learning, induces an accumulated error through the learning stage \cite{thrun}. In a function approximation setting, such an estimation noise is unavoidable as deep neural networks represent the action-value functions \cite{td3}. This inaccuracy in the action values is further exaggerated due to the temporal difference learning \cite{sutton88}. On the other hand, underestimation bias is an outcome of Q-learning variants that aim to eliminate the accumulated overestimation bias and high variance on the value estimates \cite{max_min}, \cite{td3}. Even though the standard deep Q-learning modifications are shown to overcome the overestimation bias and high variance build-up, the existence of underestimated state-action values can still degrade the performance of an RL agent by assigning low values to ``good" state-action pairs, causing suboptimal and divergent behavior.

We begin by establishing that this underestimation phenomenon is present in delayed policy gradient that utilizes a pair of critics, namely, Clipped Double Q-learning \cite{td3}, in the continuous actor-critic settings. During training, the minimum of the estimates by two critics is used to construct the target Q-value in the temporal difference learning \cite{sutton88}. Unfortunately, taking the minimum of Q-values in learning the targets produces a consistent underestimation of Q-value estimates despite the decoupled actor and critics. To address this problem, we first show that variance of the reinforcement signals that the agent encounters during the learning phase increases the underestimation bias. Then, to overcome underestimation bias and improve the performance of the deterministic actor-critic methods, we introduce a deep Q-learning variant that combines the ideas behind Clipped Double Q-learning \cite{td3}, and Maxmin Q-learning \cite{max_min}, Triplet Critic Update. Our approach leverages the notion that along with two value estimators, the value estimate of an additional estimator can be used to construct an approximate upper and lower bound to the value estimate.  

We build our modification on the state-of-the-art deterministic actor-critic algorithm, Twin Delayed Deep Deterministic Policy Gradient (TD3) \cite{td3}, and introduce the Triplet Critic Delayed Deep Deterministic Policy Gradient (TCD3) algorithm. Our deterministic actor-critic algorithm reduces the underestimation bias to a negligible margin and prevents the accumulated overestimation bias and high variance. We evaluate our algorithm on 12 continuous control tasks from OpenAI Gym \cite{gym}, where we show that Triplet Critic Update significantly improves the performance of TD3 \cite{td3}. For the reproducibility concerns, we run our experiments across a large set of seeds for the sake of a fair evaluation procedure.

\subsection{Prior Art and Comparisons}
In reinforcement learning, prior works on the function approximation error in terms of the estimation bias and high variance build-up have been studied by \cite{pendrith} and \cite{mannor}. Our work focuses on one of the outcomes of the function approximation error, namely, the underestimation bias on the action value estimates.

Several approaches reduce the effects of overestimation bias resulting from the function approximation and policy optimization in deep Q-learning. One of the successor works to the deep Q-learning \cite{dqn}, Double Q-learning \cite{ddqn}, proposed by \cite{insights_rl} and \cite{ddqn}, employs two independent action value estimators to obtain unbiased estimates of Q-values. \cite{max_min} modified the Double Q-learning \cite{ddqn} by utilizing multiple value estimators. This approach, Maxmin Q-learning \cite{max_min}, considers randomly selected value estimates of multiple critics, and a minimum of which is used in learning the Q-value target. Furthermore, methods that employ multi-step returns offer a trade-off between the variance build-up and accumulated estimation bias. Such methods are proven to be effective through distributed methods \cite{ackr}, \cite{impala}, approximate lower and upper bounds \cite{approximate_bounds}, and importance sampling \cite{precup_2001}, \cite{munos_safe_and_efficient_off_policy}. Another trade-off in these approaches is using a longer horizon rather than finding a direct solution to the accumulated error. \cite{petrik}, on the other hand, offers a solution to diminish the discount factor for reducing the contribution of each erroneous estimate.

The concern with the direct solution to the accumulated approximation error has been overcome by \cite{td3} for actor-critic settings. Their study shows the existence of overestimation bias and accumulated variance induced by the deep function approximation of Q-values. An extension of the Deep Deterministic Policy Gradient algorithm \cite{ddpg}, Twin Delayed Deep Deterministic Policy Gradient \cite{td3} which our method is built on, introduces a direct solution to the problems of the function approximation error through the utilization of two critics, delayed actor updates and target policy smoothing regularization. TD3 \cite{td3} is shown to produce state-of-the-art results by a large margin with a sufficient number of training iterations. Although the improvements introduced by \cite{td3} can tame the accumulated error, the usage of two critics induces an underestimation bias on Q-value estimates. To overcome the underestimation error, \cite{wd3} approaches the problem with a modification on the update rule for the critics in the TD3 \cite{td3}. In their approach, the target for the temporal difference learning \cite{sutton88} is computed through the weighted linear combination of the minimum and average of the two critics, Weighted Deep Deterministic Policy Gradient (WD3) \cite{wd3}. \cite{tadd} extends the WD3 \cite{wd3} by increasing the number of Q-networks to three. Both studies have shown that the weighted linear combinations of Q-networks can obtain more accurate action value estimates and higher evaluation returns.

\subsection{Contributions}
Our contributions are as follows:
\begin{enumerate}
    \item We show that if the agent receives reinforcement signals throughout the learning phase vary on a large scale, the estimation bias on the Q-value estimates severely increases.
    \item We demonstrate by empirical results that for the cases in which the overestimation and variance of the value estimates are eliminated, the underestimation bias can still degrade the performance of the RL agent in terms of the evaluation returns and convergence speed.
    \item We introduce a Double Q-learning \cite{ddqn} variant that upper and lower bounds the value estimates without any introduction of a hyper-parameter. In this way, we reduce underestimating to a negligible level and keep the overestimation and high variance build-up removed. 
    \item Through an extensive set of experiments, we show that our method improves the convergence speed and performance of the state-of-the-art on 12 challenging OpenAI Gym \cite{gym} continuous control tasks.
\end{enumerate}

\section{Background}
Reinforcement learning paradigm considers an agent interacting with its environment to learn the optimal, reward-maximizing behavior. The standard reinforcement learning can be formalized by a Markov Decision Process (MDP) defined by the tuple $(\mathcal{S, A}, p_{M}, r, \gamma)$ that involves a state space $\mathcal{S}$, an action space $\mathcal{A}$, and transition dynamics $p_{M}(s' | s, a)$ for the MDP in interest denoted by $M$. At each discrete time step $t$, given an observed state $s \in \mathcal{S}$, the agent chooses an action $a \in \mathcal{A}$ with respect to its policy $\pi: \mathcal{S} \rightarrow \mathcal{A}$ which can be either stochastic or deterministic, and receives a reward $r$ from a reward function $\mathcal{R: S \times A \rightarrow} \mathbb{R}$, and then observes a next state $s'$. The cumulative reward which the agent tries to maximize is defined as the discounted sum of rewards $R_{t} = \sum_{i = t}^{T}\gamma^{i - t}r(s_{i}, a_{i})$ where discount factor $\gamma \in [0, 1)$ is scaling long-term rewards such that short-term rewards can be prioritized more. 

Reinforcement learning aims to obtain the optimal policy $\pi^{*}_{\phi}$ parameterized by $\phi$ that maximizes the expected return $J(\phi) = \mathbb{E}_{s_{i} \sim p_{\pi}, a_{i} \sim \pi}[R_{0}]$. In continuous control domain, parametrized policies which are usually approximated by deep neural networks, can be updated by computing the gradient of the expected return $\nabla_{\phi}J(\phi)$. In an actor-critic setting, the policy $\pi$, also referred as the actor, can be updated via deterministic policy gradient (DPG) algorithm \cite{dpg}:
\begin{equation}
    \nabla_{\phi}J(\phi) = \mathbb{E}_{s \sim p_{\pi}}[\nabla_{a}Q^{\pi}(s, a) |_{a = \pi(s)} \nabla_{\phi}\pi_{\phi}(s)].
\end{equation}
The expected return when taking action $a$ after observing the state $s$ while following the policy $\pi$, $Q^{\pi}(s, a) = \mathbb{E}_{s_{i} \sim p_{\pi}, a_{i} \sim \pi}[R_{t} | s, a]$, is also called the critic or action-value function of the agent that values the quality of a state-action pair. The action-value function or the critic is used to evaluate a learning agent's current policy and improve the policy to obtain higher quality choices of actions, i.e., a better policy.

In Q-learning, when the transition probability of an environment is known, the state-value function, $Q^{\pi}$ can be estimated recursively through Bellman equation \cite{belmann} given the transition tuple $(s, a, r, s')$:
\begin{equation}
    Q^{\pi}(s, a) = r + \gamma\mathbb{E}_{s', a'}[Q^{\pi}(s', a')], \quad a' \sim \pi(s').
\end{equation}

For a large state space, the action value can be estimated by a function approximator $Q_{\theta}(s, a)$ parameterized by $\theta$. In the deep setting of Q-learning \cite{dqn}, the critic network is updated through the temporal difference learning by a secondary frozen network $Q'_{\theta}(s, a)$, also known as the target network. In this way, a fixed target $y$ can be achieved to update the critic network:
\begin{equation}
    y = r + \gamma Q_{\theta'}(s', a'), \quad a' \sim \pi_{\phi'}(s'),
\end{equation}
where the actions for the next state are chosen from a separate target actor-network $\pi_{\phi'}$ in the actor-critic setting for continuous action spaces. The target networks are either updated by a small proportion $\tau$ at each time step, $\theta' \leftarrow \tau\theta + (1 - \tau)\theta'$, called soft-update, or periodically to exactly match the current networks. Such an update rule is orthogonal to any deep reinforcement learning method that utilizes a separate target network. For instance, it can be applied to off-policy methods that sample mini-batches of transition tuples from the experience replay buffer \cite{exp_replay} for the update.

\section{The Underestimation in Deterministic Actor-Critic Methods}
\label{section:underestimation_problem}

In discrete action domains, overestimation due to the analytical maximization of action values is an evident and widely studied artifact \cite{ddqn, insights_rl, precup_2001, max_min, averaged_dqn}. For actor-critic settings, the existence and effects of the overestimation have been proven by \cite{td3} through the policy updates via gradient descent. In their approach, the minimum value of these critics is used to compute the target action value at each iteration along with delayed actor updates and smoothed target policy value. However, such utilization of the minimum operator to overcome the overestimation of the Q-values may introduce an underestimation bias \cite{td3}, \cite{wd3}, \cite{tadd}. We begin by proving through basic assumptions and statements that the underestimation phenomenon occurs in continuous control, actor-critic methods in environments with different reinforcement signals. Then, we introduce our modified target Q-value update rule to reduce the underestimation bias while staying in the ``safe zone" of function approximation error in the actor-critic setting.

In the TD3 algorithm \cite{td3}, the policy is updated with respect to the value estimates of one of two approximate critics. Without loss of generality, we assume that the policy is updated with respect to the first approximate critic, which is denoted by $Q_{\theta_{1}}(s, a)$ through the deterministic policy gradient. We show that in an environment with rewards that vary on a large scale, the target update rule for the Q-values induces an underestimation bias. 

Let $\phi_{\mathrm{approx}}$ define the parameters from the actor update by the maximization of the first approximate critic $Q_{\theta_{1}}(s, a)$:
\begin{equation}
\text{\footnotesize$
    \phi_{\mathrm{approx}} = \phi + \frac{\alpha}{Z_{1}}\mathbb{E}_{s \sim p_{\pi}}[\nabla_{\phi}\pi_{\phi}(s)\nabla_{a}Q_{\theta_{1}}(s, a)|_{a = \pi_{\phi}(s)}],$\normalsize}
\end{equation}
\normalsize
where $Z_{1}$ is the gradient normalizing term such that $Z^{-1}||\mathbb{E}[\cdot]|| = 1$. As the actor is optimized with respect to $Q_{\theta_{1}}(s, a)$ and the gradient direction is a local maximizer, there exists $\xi$ sufficiently small such that if $\alpha < \xi$, then the \textit{approximate} value of $\pi_{\mathrm{approx}}$ by the first critic will be bounded below by the \textit{approximate} value of $\pi_{\mathrm{approx}}$ by the second critic:
\begin{equation}\label{eq:td3_like_underestimation}
    \mathbb{E}[Q_{\theta_{1}}(s, \pi_{\mathrm{approx}}(s))] \geq \mathbb{E}[Q_{\theta_{2}}(s, \pi_{\mathrm{approx}}(s))].
\end{equation}
Note that in the latter equation, the estimated action values by both critics are overestimated. Then, we can treat the function approximation error as a Gaussian random variable:
\begin{align}\label{eq:error_gaussian_def}
\begin{split}
    Q_{\theta_{1}}(s, a) - Q^{*}(s, a) &= G_{1} \sim \mathcal{N}(\epsilon_{1}, \sigma_{1}), \\
    Q_{\theta_{2}}(s, a) - Q^{*}(s, a) &= G_{2} \sim \mathcal{N}(\epsilon_{2}, \sigma_{2}).
\end{split}
\end{align}
By (\ref{eq:td3_like_underestimation}) and (\ref{eq:error_gaussian_def}), we have $\epsilon_{1} \geq\epsilon_{2} \geq 0$. Moreover, as the presence of the delayed actor updates, the mean function approximation errors by both critics are not very distant due to the decoupling the actor and first critic, i.e., $\epsilon_{1} - \epsilon_{2} \approx 0$. Then, by the first moments of the minimum of two correlated Gaussian random variables \cite{nadaraj}, the expected function approximation error for the Clipped Double Q-Learning algorithm \cite{td3} becomes:
\begin{equation}
    \mathbb{E}[\underset{i=1, 2}{\mathrm{min}}\{G_{i}\}] = \frac{\epsilon_{1} + \epsilon_{2}}{2} - \frac{\theta}{\sqrt{2\pi}},
\end{equation}
where $\theta \coloneqq \sqrt{\sigma_{1}^{2} + \sigma_{2}^{2} - 2\rho\sigma_{1}\sigma_{2}}$ and $\rho$ is the correlation coefficient between the error distributions, $G_{1}$ and $G_{2}$. The error Gaussian's are correlated since the critics are not entirely independent due to the use of the same experience replay buffer and opposite critics in learning the approximate targets \cite{td3}. If $\sigma_{1}, \sigma_{2} > \sqrt{\frac{\pi}{1 - \rho}}\epsilon_{1}$, then the action value estimate will be underestimated:
\begin{equation}
    \mathbb{E}[\underset{i=1, 2}{\mathrm{min}}\{Q_{\theta_{i}}(s, a)\} - Q^{*}(s, a)] < 0.
\end{equation}

It can be observed from the underestimation condition that for a highly correlated pair of critics, underestimation does not exist. However, because of the delayed policy updates, the correlation between the critics is regularized \cite{td3}. Thus, a weak or moderate correlation between the pair of critics is expected \cite{td3}, which increases the possibility of underestimation.

Although delayed policy updates and the minimization of the value estimates aim to reduce error growth, the variances of the value estimates are not eliminated since they are proportional to the variance of the future rewards and estimation errors \cite{td3}. In function approximation setting, the Bellman equation is never exactly satisfied, yielding erroneous value estimates as a function of the true TD-error \cite{off_policy}, as expressed by (\ref{eq:error_gaussian_def}). Then, it can be shown that the variance of the value estimates overgrow as the agent interacts with the environment and observes varying reward signals \cite{sutton2018reinforcement}. We can express the approximate Q-values in terms of the expected value of the discounted cumulative rewards, as shown in \cite{td3}:
\begin{align}
\begin{split}
\text{\small$
    Q_{\theta_{i}}(s, a) = \mathbb{E}_{s_{i} \sim p_{\pi}, a_{i} \sim \pi}[\sum_{i = t}^{T}\gamma^{i - t}r_{i}] + \epsilon_{i}\sum_{i = t}^{T}\gamma^{i - t}.$\normalsize}
\end{split}
\end{align}
Suppose the expected value of the estimation error is constant for both critics. In that case, varying reinforcement signals increase the variance of the value estimates, which results in an underestimation error. The agent extensively explores the environment, a mandatory requirement for continuous action domain \cite{off_policy}, the variance of the reward signals that the agent encounters start growing, and underestimation bias will become inevitable. Furthermore, the minimum operator eliminates the accumulating error due to the temporal difference learning, and thus, underestimation is far more preferable to overestimation bias in actor-critic setting \cite{td3}. Nevertheless, underestimating a value estimate may discourage the agent from choosing good state-action pairs for an extended period and reinforce the agent to value suboptimal state-action pairs more. 
\begin{figure*}[htbp]
	\centering
	
	\subfigure[Ant-v2]{
		\includegraphics[width=2in, keepaspectratio]{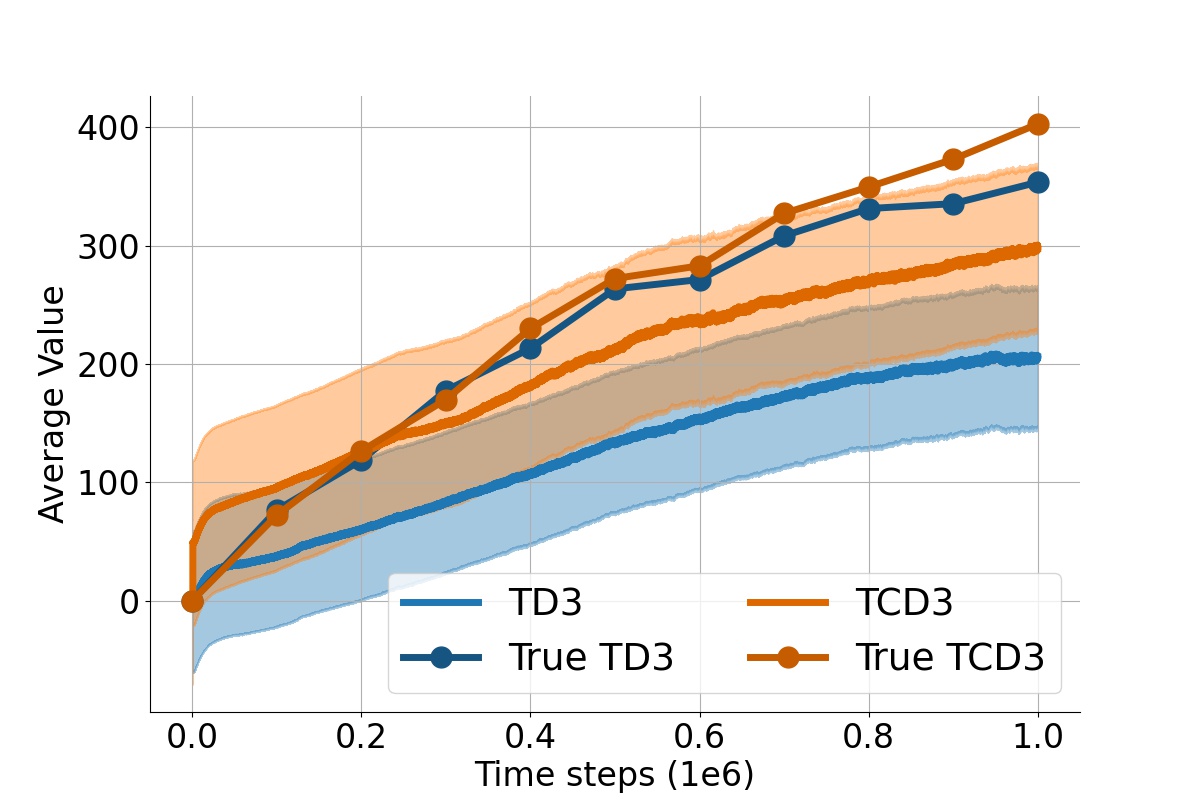}
	}
	\hspace{-0.3in}%
	\vspace{-0.06in}%
	\subfigure[BipedalWalker-v3]{
		\includegraphics[width=2in, keepaspectratio]{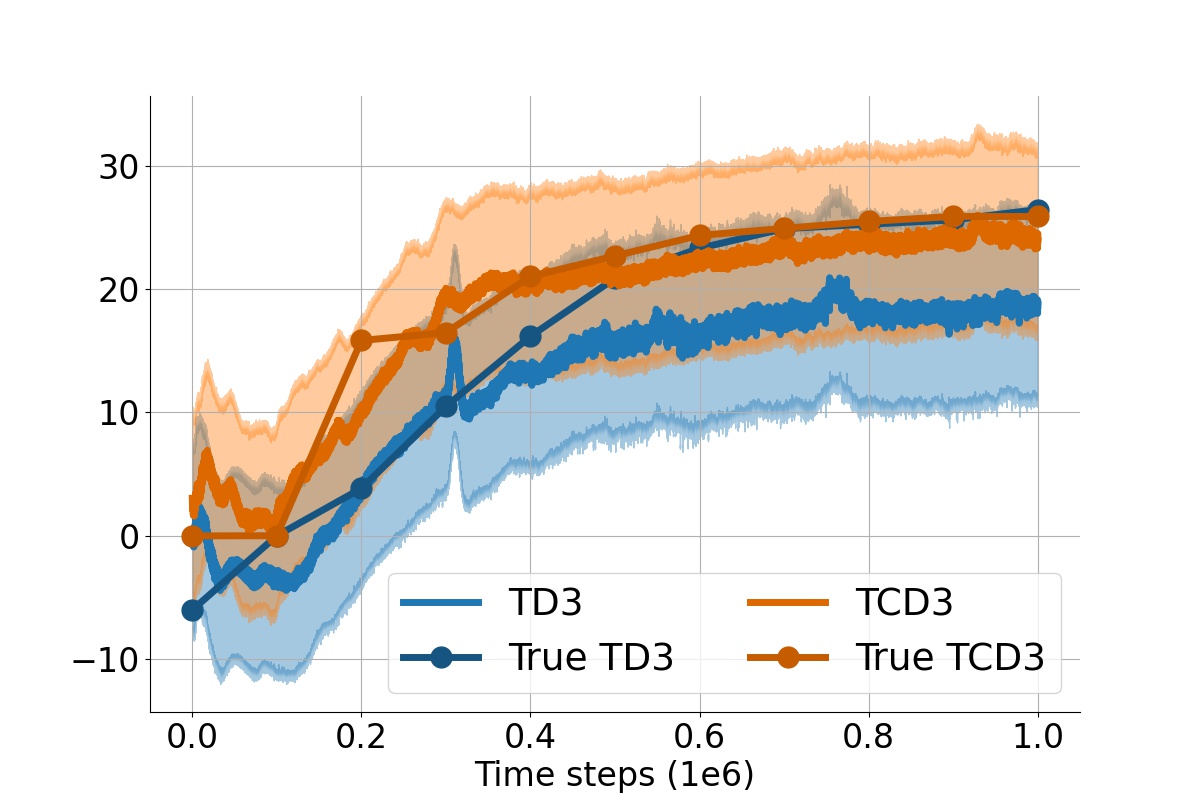}
	}
	\hspace{-0.3in}%
	\vspace{-0.06in}%
	\subfigure[Humanoid-v2]{
		\includegraphics[width=2in, keepaspectratio]{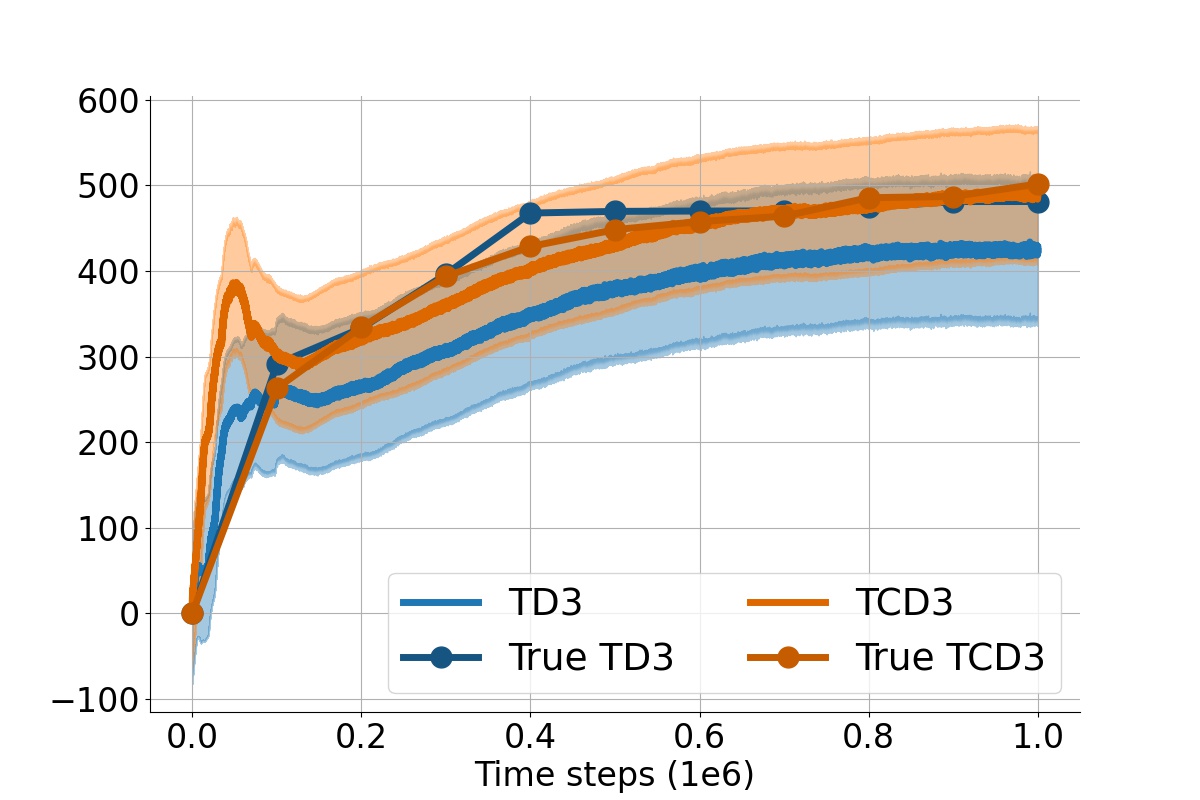}
	}
	\hspace{-0.3in}%
	\vspace{-0.06in}%
	\subfigure[HumanoidStandup-v2]{
		\includegraphics[width=2in, keepaspectratio]{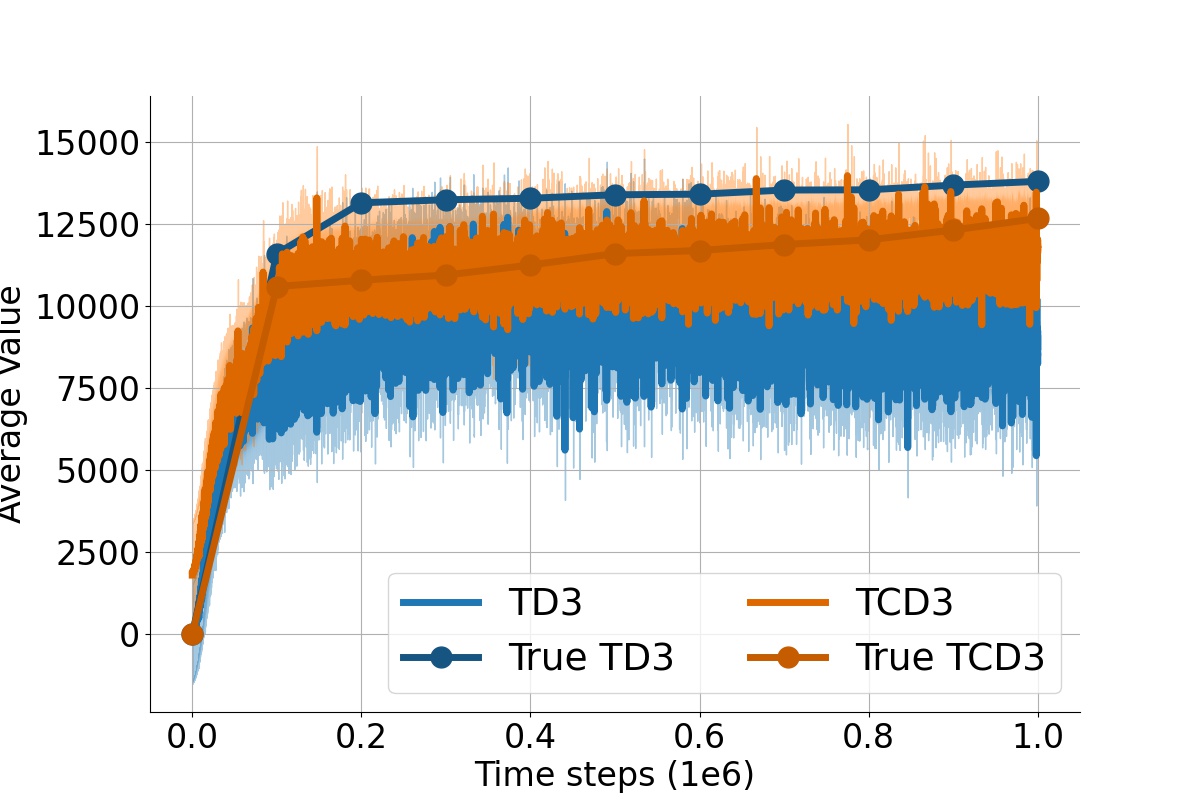}
	}
	\hspace{-0.3in}%
	\vspace{0.04in}%
	\subfigure[LunarLanderContinuous-v2]{
		\includegraphics[width=2in, keepaspectratio]{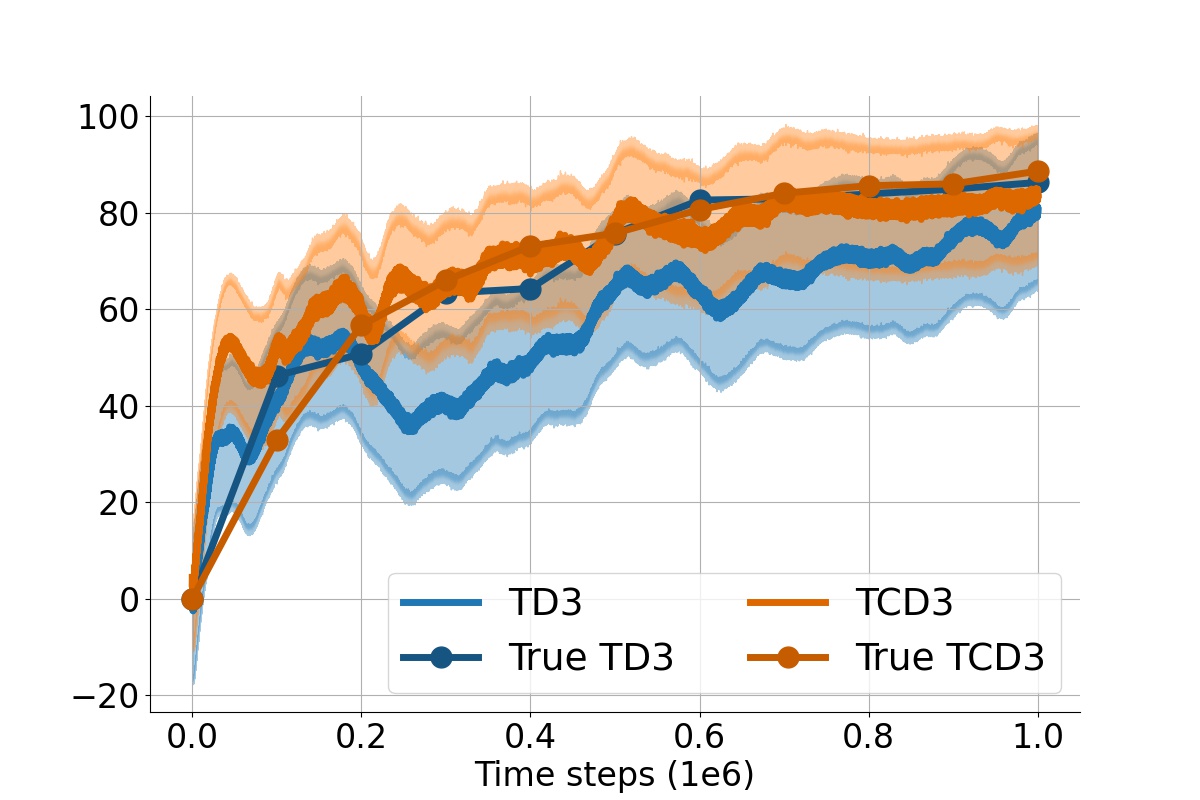}
	}
	\hspace{-0.3in}%
	\vspace{0.04in}%
	\subfigure[Swimmer-v2]{
		\includegraphics[width=2in, keepaspectratio]{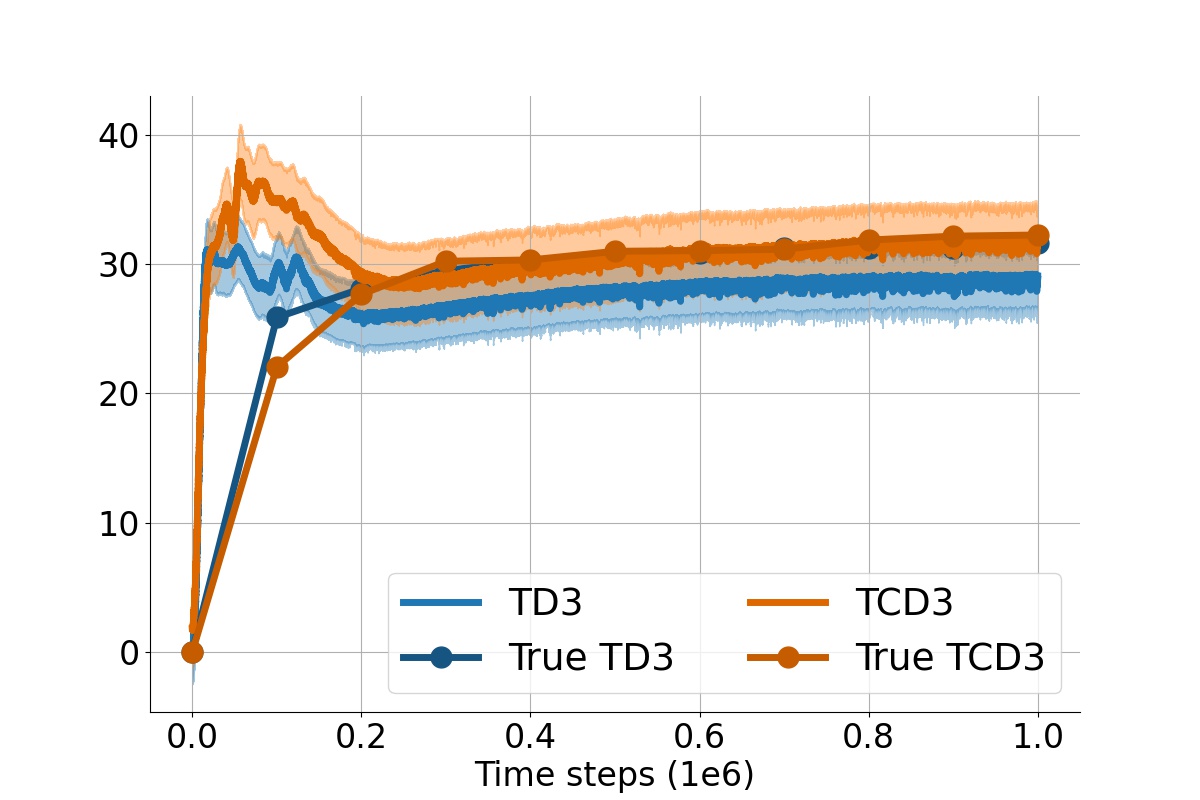}
	}
	
	\caption{Measuring estimation bias of fine-tuned TD3 versus TCD3 while learning on MuJoCo and Box2D environments over 1 million time steps. Estimated and true Q-values are computed through Monte Carlo simulation for 1000 samples.} 
	\label{td3_tcd3_q_estimation}
\end{figure*}

\subsection{Does the theoretical underestimation due to the minimum operator occur in practice?}
This question can be answered by observing value estimates produced by the target Q-value update with the minimum of two critics over a training duration along with the actual Q-values while the agent is learning on a set of OpenAI Gym \cite{gym} continuous control tasks. The true Q-values are estimated through the average discounted sum of rewards over randomly initialized 1000 episodes following the current policy. Estimated Q-values are also computed using the average Q-value estimates by the critics on the 1000 sampled episodes.

In Fig. \ref{td3_tcd3_q_estimation}, there exists an apparent underestimation bias that occurs on the learning procedure. The underestimation bias keeps growing or follows a pattern parallel with the actual values depending on the environment. These empirical findings verify our claims; that is, the approximate critics start by overestimating the actual Q-values. However, after a while in which the agent starts an extensive exploration procedure, the variances of the approximate critics increase, and an unavoidable underestimation arises. Moreover, for some environments such as Ant, as the actual state-action values increase due to the varying rewards, the variances of the approximate critics increase, and the underestimation on the value estimates keeps growing even with the delayed target and actor updates. Whereas, when the true Q-values reach a steady-state, the underestimation settles to a certain level which can also be observed from Humanoid, Swimmer, and BipedalWalker environments. Even though in the set of OpenAI Gym \cite{gym} tasks, the continuous state and multi-dimensional action spaces are contributing factors to the growth in the variance of the action-values, the scale of these tasks is still tiny compared to the real-world settings \cite{off_policy}. Therefore, underestimating value estimates would be a more detrimental and inevitable problem for tasks with larger-scaled state and action spaces.  

To remedy the observed underestimation bias, we next introduce our novel modification on the target Q-value update, Triplet Critic Update, which significantly reduces the underestimation bias by the minimum of two approximate critics. 

\section{Triplet Critic Update}
Several approaches have been proposed to overcome the underestimation bias introduced by the minimum of two critics. These approaches differ in terms of the treatments of the function approximation error as random variables. Although these proposals can overcome the underestimation bias introduced by the target Q-value update, they introduce additional hyper-parameters to be tuned. This section presents our parameter-free, novel modification on the target update rule for deterministic actor-critic methods for continuous control, which is also a variant of Double Q-learning \cite{ddqn} but with three critics.

In the current approach to overcome the overestimated action values, we highlight that the bias introduced by the approximate critics is a function of the expected function approximation error by the critics and variance of the value estimates. As the expected function approximation error is assumed to be constant for both critics, depending on the variance of the value estimates, which is proportional to the encountered reinforcement signals, the function approximation error can either be an underestimation or overestimation. In practice, however, we observe that the variance of the value estimates is much greater than the expected function approximation error as the continuous control tasks require a comprehensive exploration procedure. Hence, an underestimation error occurs caused by the objective computation for the action value functions that employ the minimum operator as in TD3 \cite{td3}. As a result, the critic does not value good state-action pairs as much as they should be. 
To address this problem that has been approached from different perspectives, we propose to simply upper and lower-bound the biased value estimates of three critics without introducing any additional hyper-parameter. This results in taking the minimum of the maximum of two critics and a single distinct critic, which gives the modified target update rule, Triplet Critic Update:
\begin{equation}
\text{\small$
    y = r + \gamma \mathrm{min}\left(\underset{i = 1, 2}{\mathrm{max}}(Q_{\theta'_{i}}(s', \pi_{\phi'}(s')), Q_{\theta'_{3}}(s', \pi_{\phi'}(s')\right).$\normalsize}
\end{equation}
As the actor is optimized with respect to the first critic, the second and third critics can be represented by the same probability distribution, i.e., $G_{3} \sim \mathcal{N}(\epsilon_{2}, \sigma_{2})$. Under the same assumptions presented in the previous section, the expected function approximation error for the Triplet Critic Update can be computed by the extensions of \cite{nadaraj} and \cite{expect_of_moments_of_max}:
\begin{align}
\begin{split}
    \text{\small$\mathbb{E}[\mathrm{min}(\mathrm{max}(G_{1}, G_{2}), G_{3})]$} &= \text{\small$(\epsilon_{1} + 3\epsilon_{2}) / 4 - \theta / 2\sqrt{2\pi}$}, \\ &= \text{\small $\frac{(\mathbb{E}[\underset{i=1, 2}{\mathrm{min}}\{G_{i}\}] + \epsilon_{2})}{2}$\normalsize}.
\end{split}
\end{align}
This expected error value is the average of the biases introduced by the minimum of two target action values as in TD3 \cite{td3}, and an approximate critic as in DDPG \cite{ddpg}. If the overestimation is relatively larger than the underestimation, there will be a slight overestimation by this update rule, and vice versa for a relatively larger underestimation. Regardless, such slight estimation errors can be tolerated by the agent compared to TD3 \cite{td3} and DDPG \cite{ddpg}. However, in practice, the variance of the value estimates is usually large, and hence, the Triplet Critic Update will underestimate the value estimates. Nonetheless, the bias will be slightly larger than half of the bias in TD3 \cite{td3} which will greatly reduce the underestimation and obtain more accurate value estimates without introducing additional hyper-parameter to be tuned. In addition, as the underestimation is preferable to the overestimation due to non-accumulating error characteristics, \cite{td3}, the Triplet Critic Update remains in the ``safe zone" of the function approximation in general, by keeping the overestimation and accumulated variance eliminated. Finally, it is observable in our simulations that for the extreme cases in which the variance of the value estimates is very large or very small, the Triplet Critic Update offers more accurate estimates than TD3 \cite{td3} for both ends. 

\begin{algorithm}[htbp]
    \caption{TripletCriticUpdate}
    \begin{algorithmic}
    \small
        \STATE \textbf{Input} $Q_{\theta_{1}'}, Q_{\theta_{2}'}, Q_{\theta_{3}'}$, $s'$, $\tilde{a}$
        \STATE $y \leftarrow r + \gamma \mathrm{min}(\mathrm{max}(Q_{\theta'_{1}}(s', \tilde{a}), Q_{\theta'_{2}}(s', \tilde{a})), Q_{\theta'_{3}}(s', \tilde{a}))$
        \RETURN $y$
    \end{algorithmic}
    \normalsize
    \label{alg:triplet critic update}
\end{algorithm}
\begin{algorithm}[hb]
    \caption{TCD3}
    \begin{algorithmic}
    \small
        \STATE Initialize critic networks $Q_{\theta_{1}}, Q_{\theta_{2}}, Q_{\theta_{3}}$, and actor network $\pi_{\phi}$ with randomly initialized parameters $\theta_{1}, \theta_{2}, \theta_{3}, \phi$
        \STATE Initialize target networks $\theta_{i}' \leftarrow \theta_{i}, \phi' \leftarrow \phi$
        \STATE Initialize replay buffer $\mathcal{B}$
        \FOR{$t = 1$ \textbf{to} $T$}
            \STATE Select action with exploration noise $a \sim \pi_{\phi}(s) + \eta$, $\eta \sim \mathcal{N}(0, \sigma)$ and observe reward $r$ and new state $s'$
            \STATE Store transition tuple $(s, a, r, s')$ in $\mathcal{B}$
            \STATE Sample mini-batch of $N$ transitions $(s, a, r, s')$ from $\mathcal{B}$
            \STATE $\tilde{a} \leftarrow \pi_{\phi'}(s') + \eta; \eta \sim \mathrm{clip}(\mathcal{N}(0, \tilde{\sigma}), -c, c)$
            \STATE $y \leftarrow $ \textbf{TripletCriticUpdate}$(Q_{\theta'_{1}}, Q_{\theta'_{2}}, Q_{\theta'_{3}}, s', \tilde{a})$
            \STATE Update critics $\theta_{i} \leftarrow \mathrm{argmin}_{\theta_{i}}N^{-1}\sum(y - Q_{\theta_{i}}(s, a))^{2}$
            \IF{$t$ mod $d$}
                \STATE Update $\phi$ by the deterministic policy gradient:
                \STATE $\nabla_{\phi}J(\phi) = N^{-1}\sum\nabla_{a}Q_{\theta_{1}}(s, a)\vert_{a = \pi_{\phi(s)}}\nabla_{\phi}\pi_{\phi}(s)$
                
                \STATE Update target networks:
                \STATE $\theta_{i}' \leftarrow \tau\theta_{i} + (1 - \tau)\theta_{i}'$
                \STATE $\phi' \leftarrow \tau\phi + (1 - \tau)\phi'$
                
            \ENDIF
        \ENDFOR
    \normalsize
    \end{algorithmic}
    \label{alg:tcd3}
\end{algorithm}

In the implementation, for the computational efficiency, the actor again should be optimized with respect to the first critic, and the utilization of more than three critics should be avoided. Therefore, the order of the critics in the nested Triplet Critic Update must remain constant. Otherwise, instability may occur if the critic with respect to which the actor is optimized and the order of critics in the update rule is altered throughout the learning phase. However, the utilization of three critics introduces additional computational complexity. If the clipped Double Q-Learning algorithm \cite{td3} is assumed to have a runtime of $\mathcal{O}(n)$, then our approach would run on $2\mathcal{O}(n)$ which increases linearly. Thus, the runtime can be reduced to $2\mathcal{O}(n) \approx \mathcal{O}(n)$ and neglected if the number of training time steps is not very large. Consequently, our modification comprehensively considers computational complexity and estimation error accuracy trade-off by introducing only a single additional critic without any hyper-parameter that requires further tuning.

We now introduce our approach built on the TD3 algorithm \cite{td3} changing the target Q-value update. Our algorithm called Triplet Critic Deep Deterministic Policy Gradient (TCD3) is summarized in Algorithm \ref{alg:tcd3}. In the next section, we present the experimental results for our algorithm in terms of both the Q-value estimation comparisons with TD3 \cite{td3}, and evaluations on several OpenAI Gym \cite{gym} continuous control environments.

\section{Experiments}
To evaluate our estimation error correction approach, we first demonstrate the estimated and actual Q-value curves for TCD3, and state-of-the-art off-policy continuous control algorithm TD3 on both MuJoCo \cite{mujoco}, and Box2D \cite{box2d} continuous control tasks interfaced by OpenAI Gym \cite{gym}. Subsequently, we evaluate and compare the learning RL agents under TCD3 and TD3 algorithms on the extended set of OpenAI Gym \cite{gym} control tasks. To provide reproducibility and allow a fair comparison, we directly use the same set of tasks from \cite{gym} with no modifications to the environment dynamics.

\begin{figure*}[t]
	\centering
    \textcolor{custom_blue}{$\blacksquare$} Fine-tuned TD3 \quad \quad \textcolor{custom_orange}{$\blacksquare$} TCD3
    
	\subfigure[Ant-v2]{
		\includegraphics[width=1.7in, keepaspectratio]{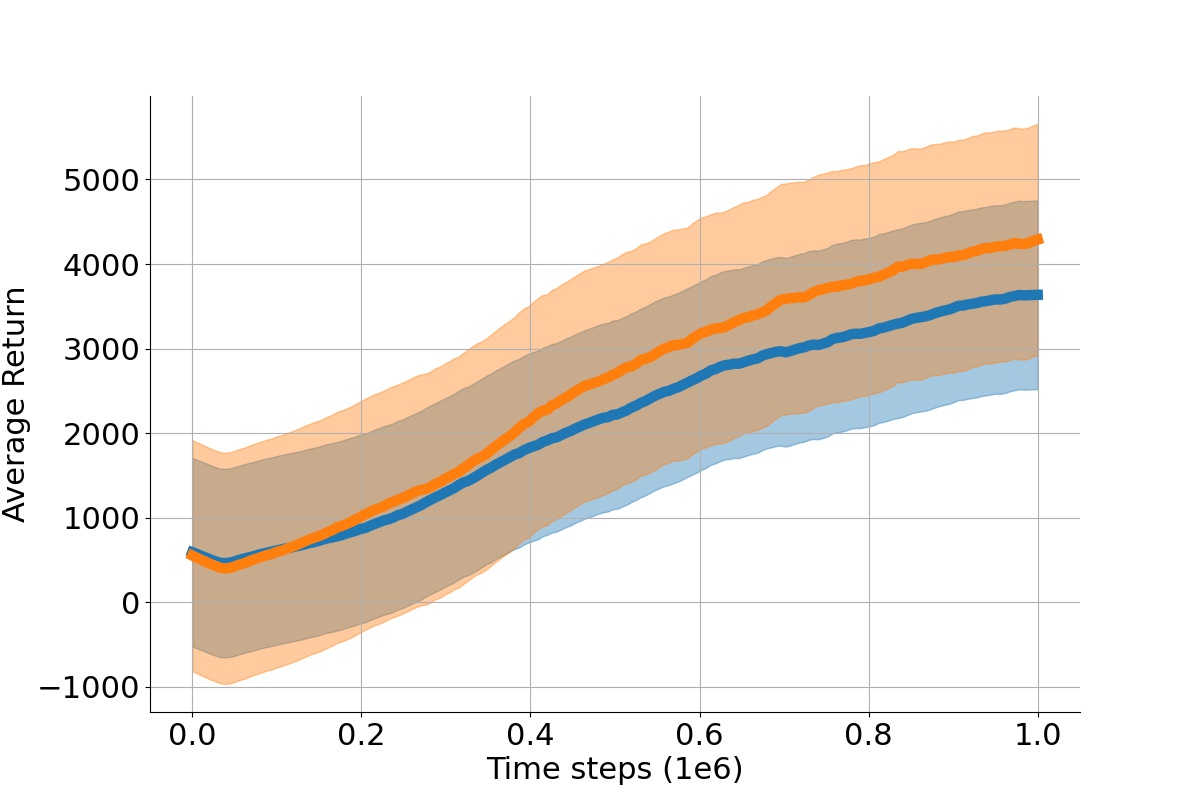}
	}
	\hspace{-0.3in}%
	\vspace{-0.04in}%
	\subfigure[BipedalWalker-v3]{
		\includegraphics[width=1.7in, keepaspectratio]{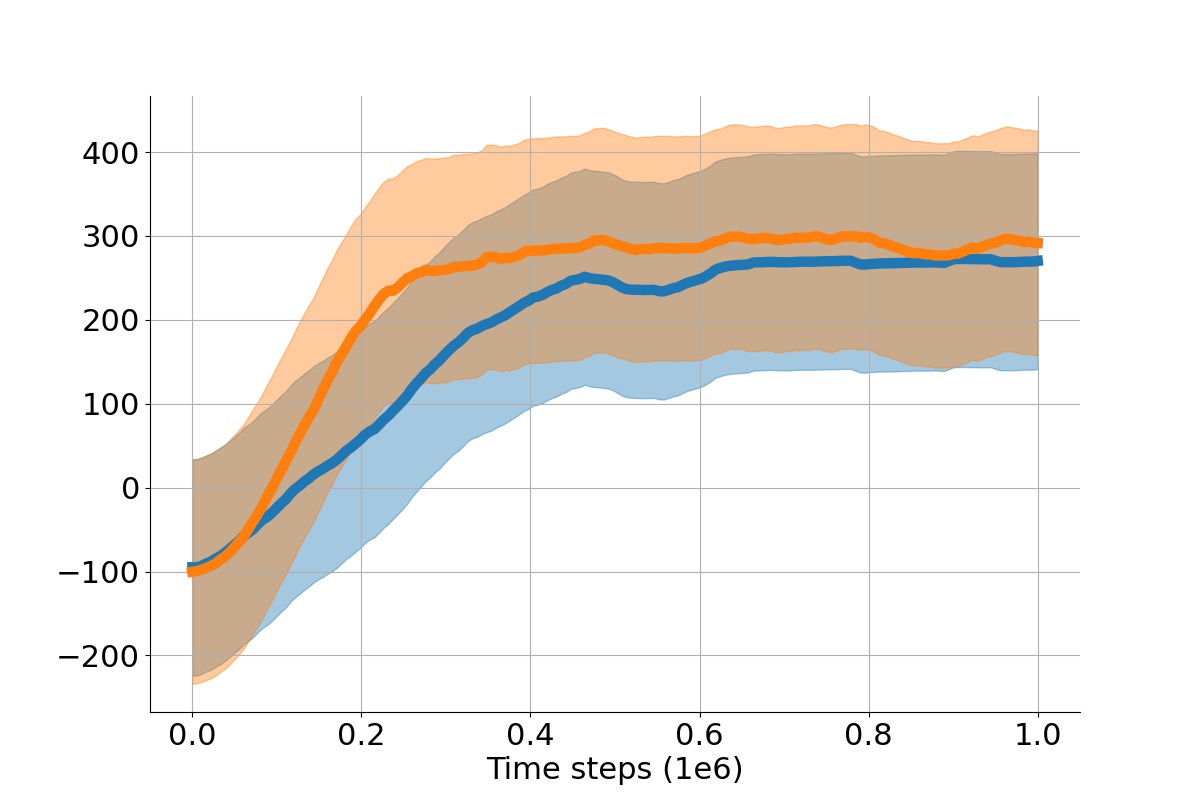}
	}
	\hspace{-0.3in}%
	\vspace{-0.04in}%
	\subfigure[HalfCheetah-v2]{
		\includegraphics[width=1.7in, keepaspectratio]{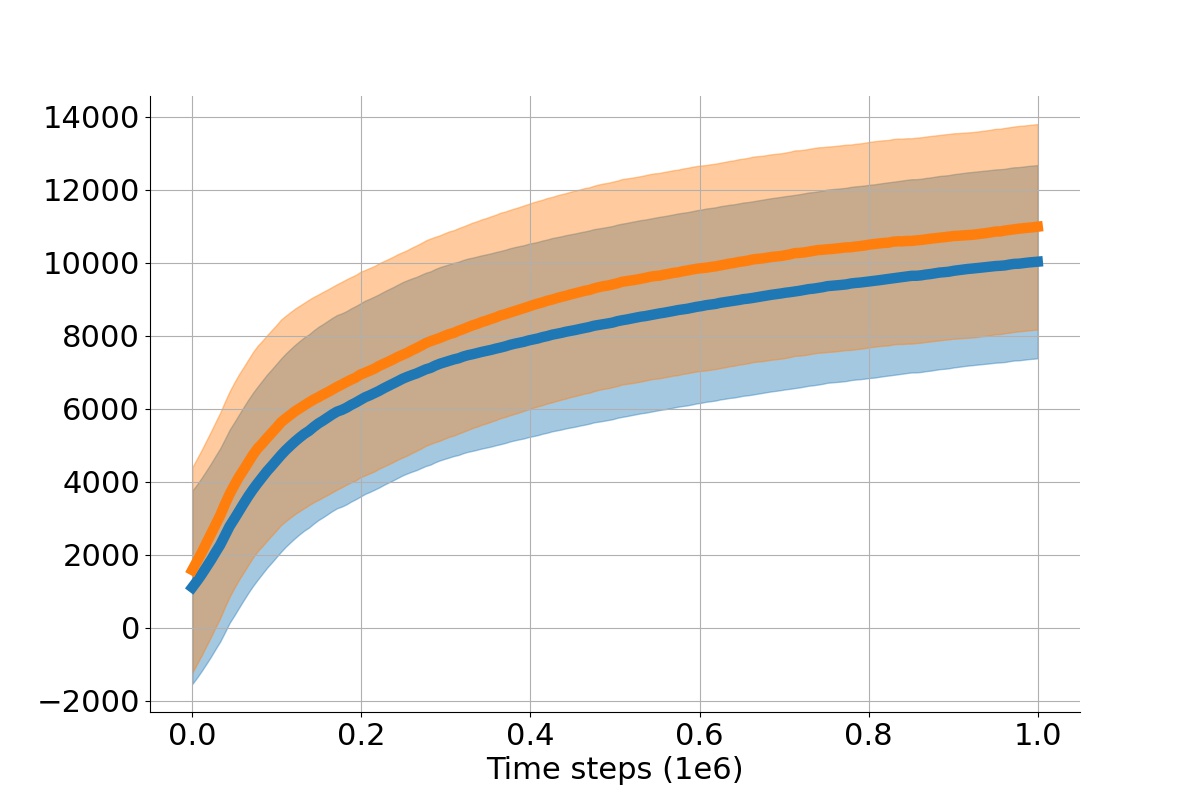}
	}
	\hspace{-0.3in}%
	\vspace{-0.04in}%
	\subfigure[Hopper-v2]{
		\includegraphics[width=1.7in, keepaspectratio]{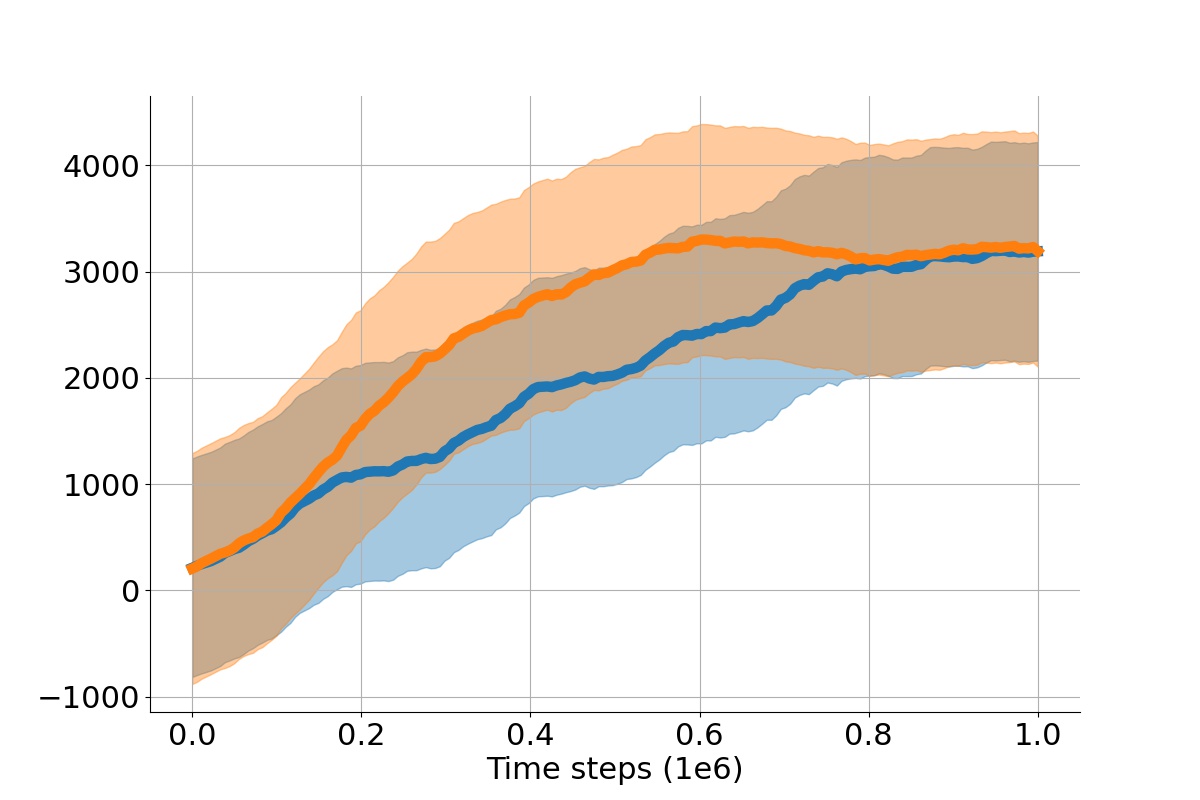}
	}
	\hspace{-0.1in}%
	\vspace{-0.04in}%
	\subfigure[Humanoid-v2]{
		\includegraphics[width=1.7in, keepaspectratio]{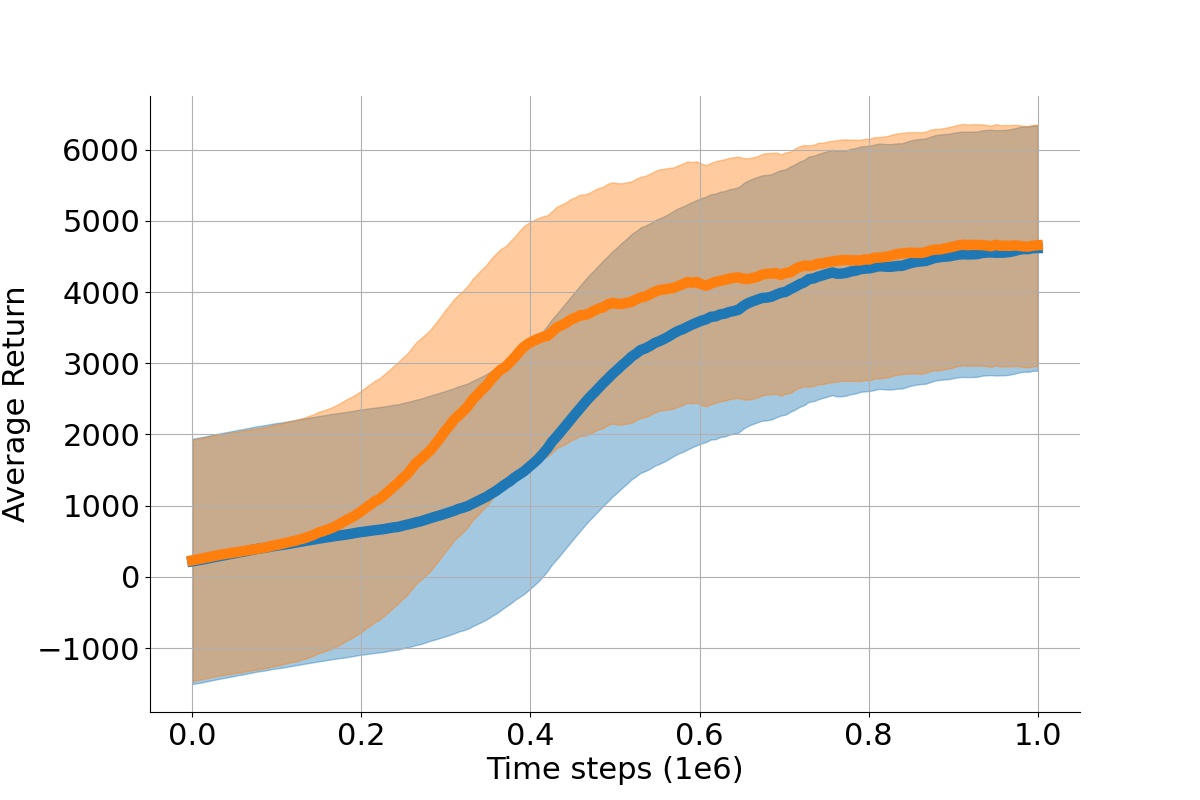}
	}
	\hspace{-0.3in}%
	\vspace{-0.03in}%
	\subfigure[HumanoidStandup-v2]{
		\includegraphics[width=1.7in, keepaspectratio]{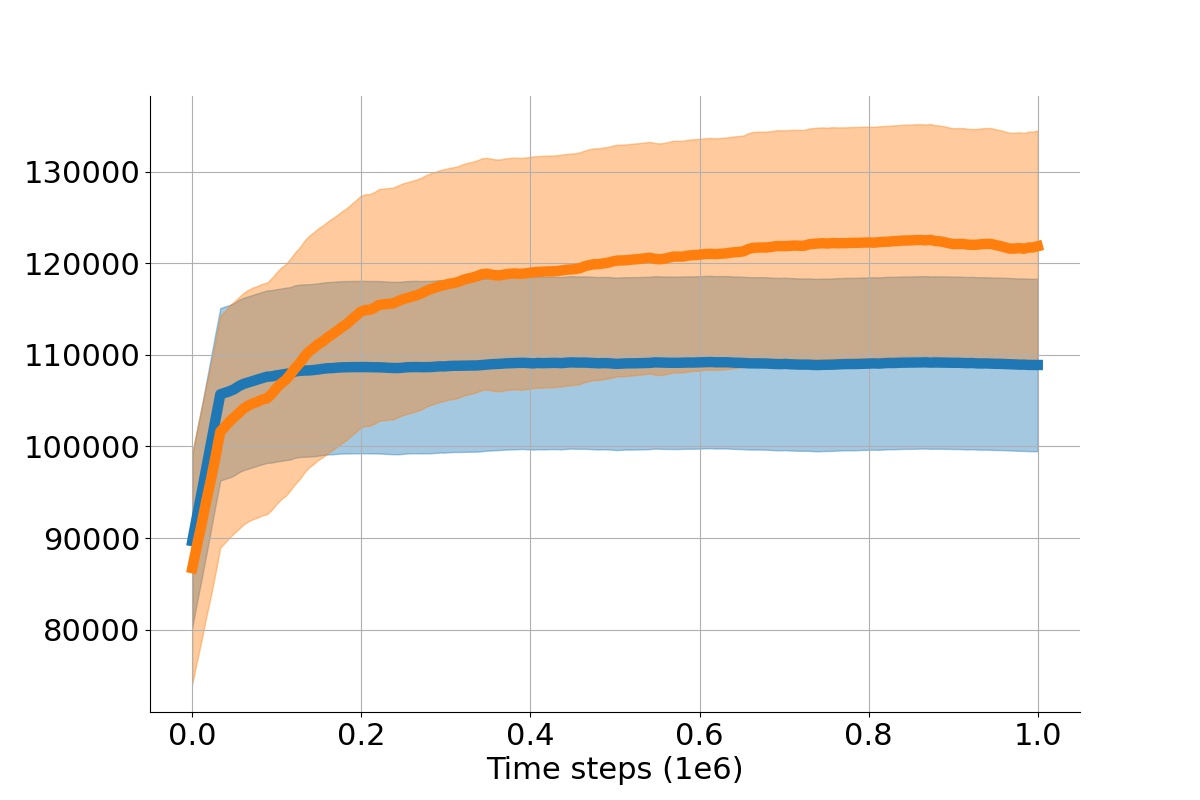}
	}
	\hspace{-0.3in}%
	\vspace{-0.03in}%
	\subfigure[InvertedDoublePendulum-v2]{
		\includegraphics[width=1.7in, keepaspectratio]{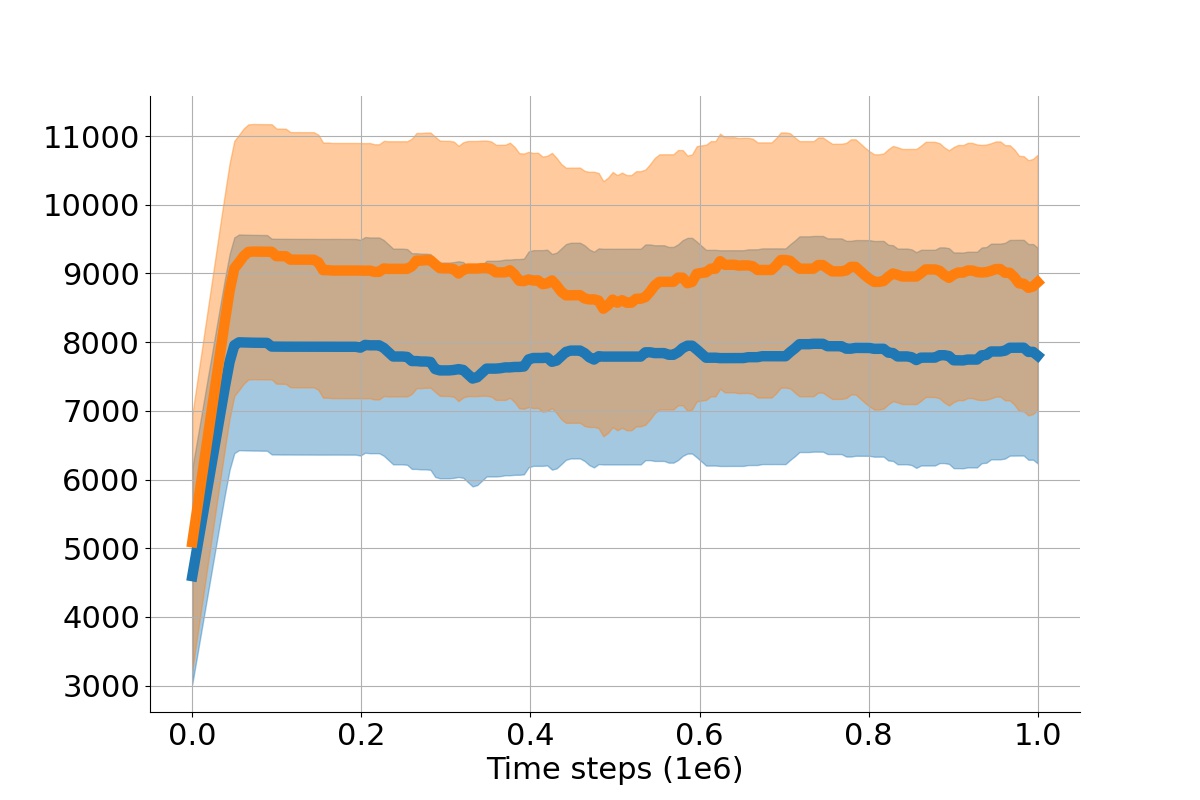}
	}
	\hspace{-0.3in}%
	\vspace{-0.03in}%
	\subfigure[InvertedPendulum-v2]{
		\includegraphics[width=1.7in, keepaspectratio]{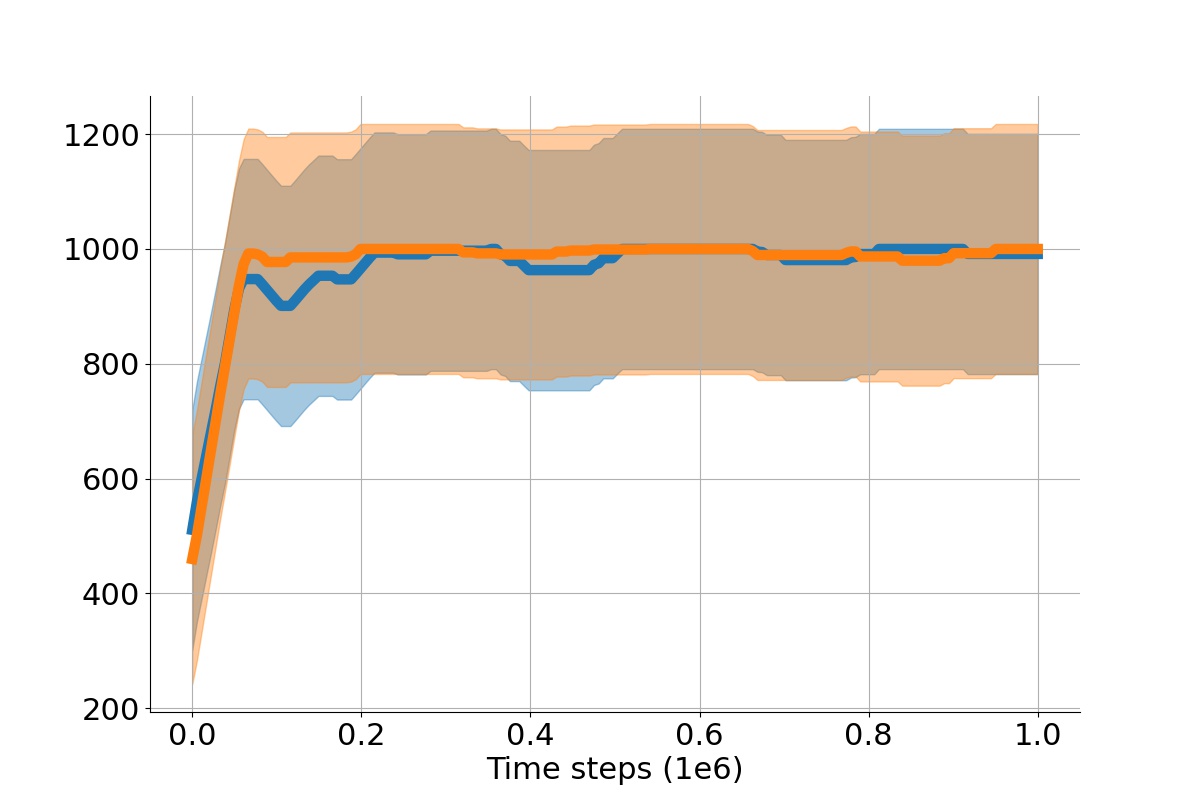}
	}
	\hspace{-0.1in}%
	\vspace{-0.03in}%
	\subfigure[LunarLanderContinuous-v2]{
		\includegraphics[width=1.7in, keepaspectratio]{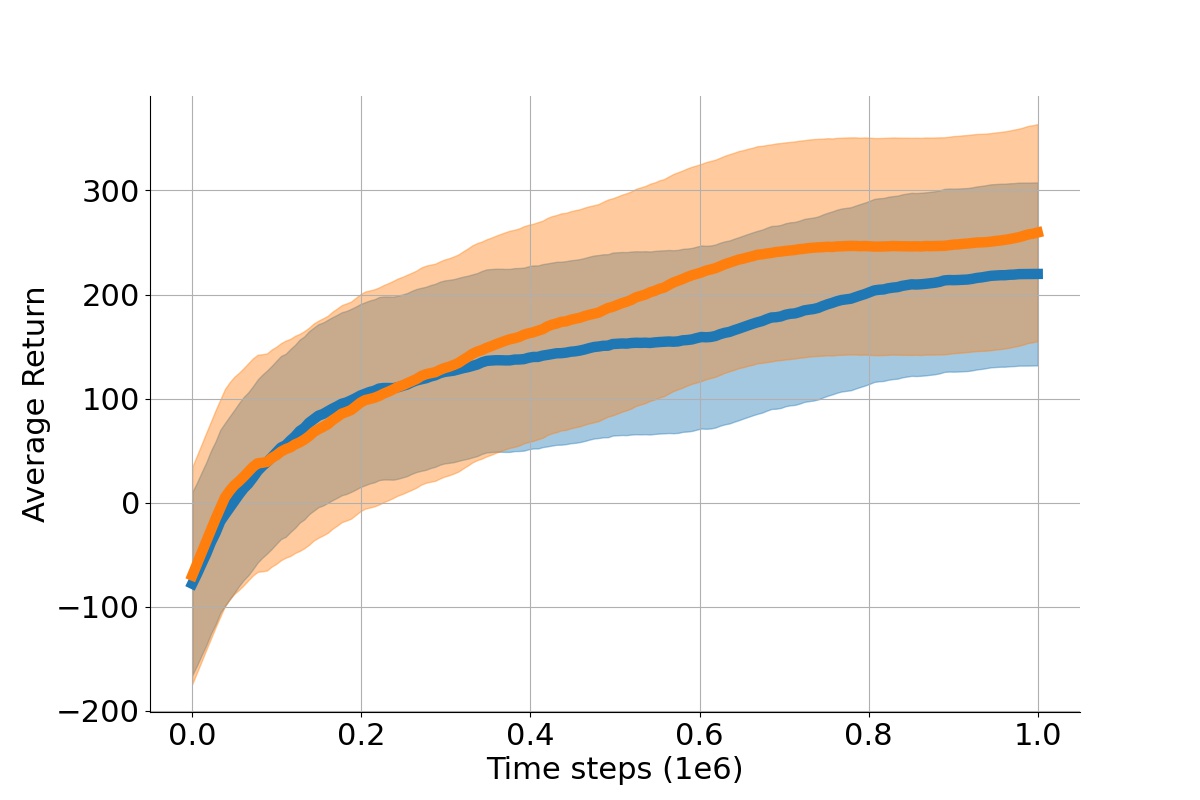}
	}
	\hspace{-0.3in}%
	\vspace{-0.04in}%
	\subfigure[Reacher-v2]{
		\includegraphics[width=1.7in, keepaspectratio]{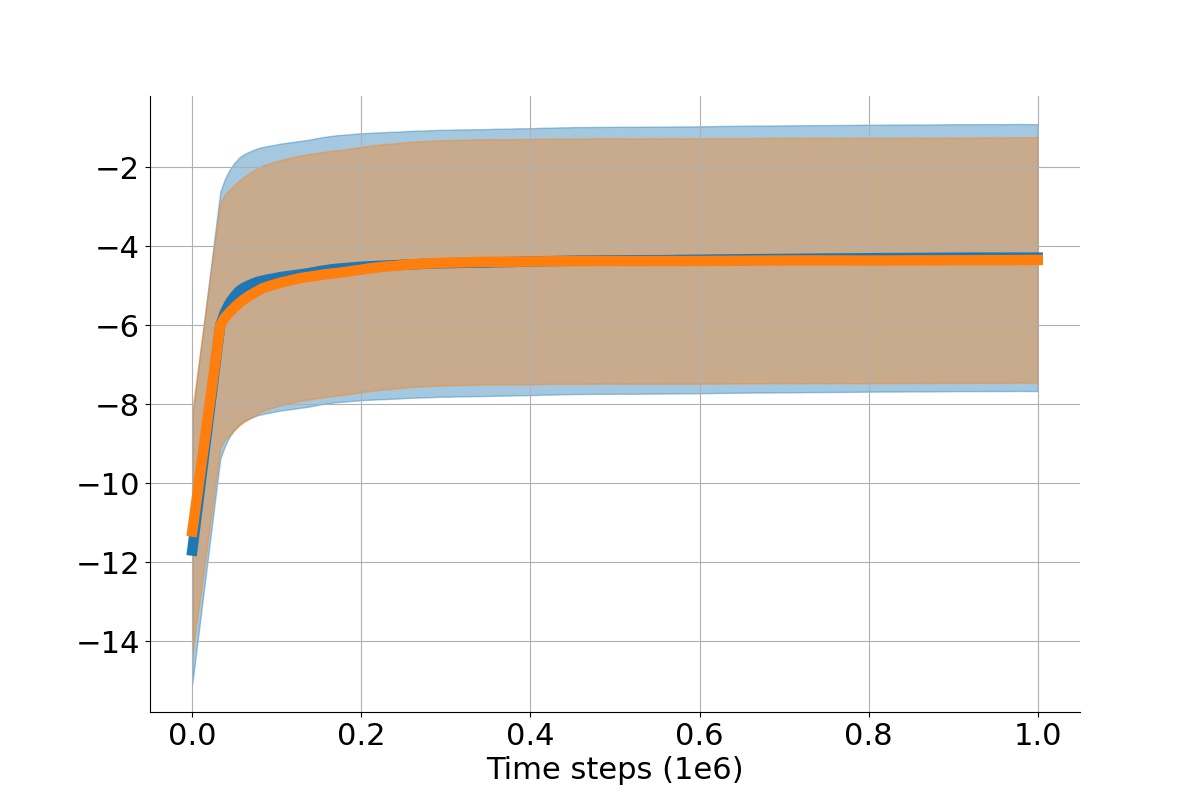}
	}
	\hspace{-0.3in}%
	\vspace{0.04in}%
	\subfigure[Swimmer-v2]{
		\includegraphics[width=1.7in, keepaspectratio]{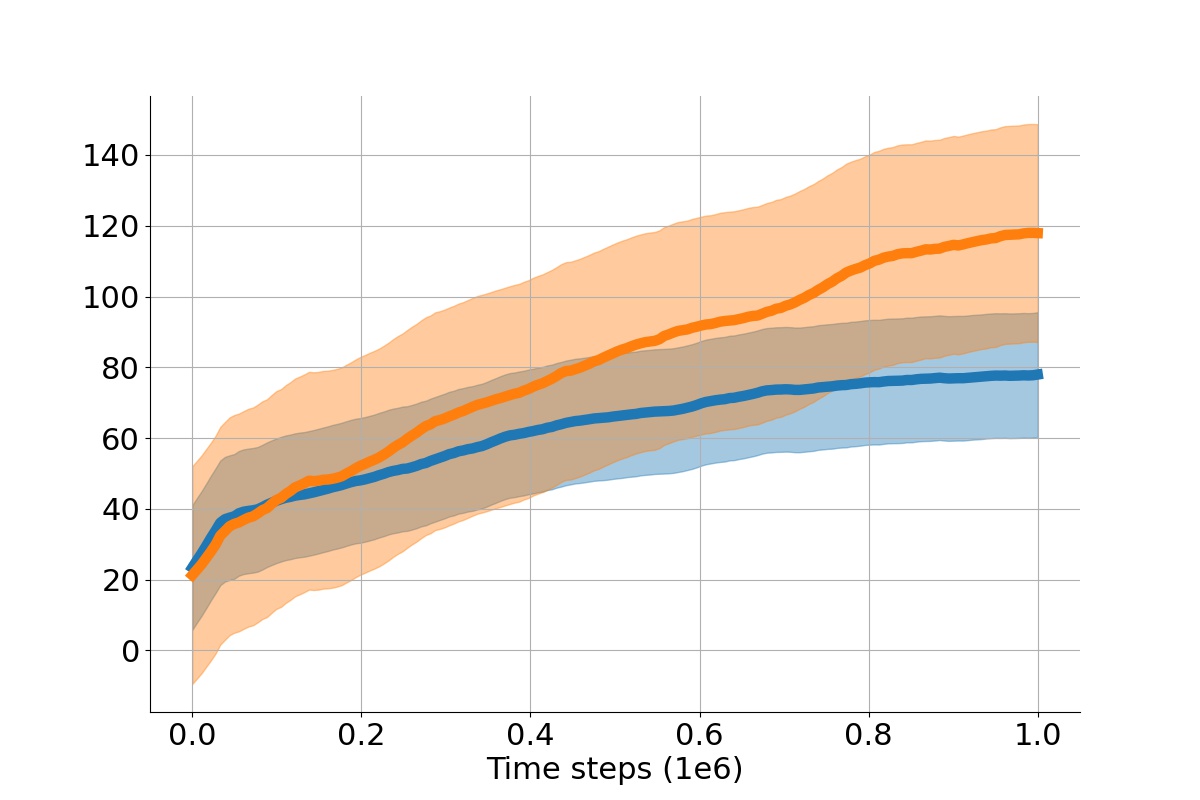}
	}
	\hspace{-0.3in}%
	\vspace{0.04in}%
	\subfigure[Walker2d-v2]{
		\includegraphics[width=1.7in, keepaspectratio]{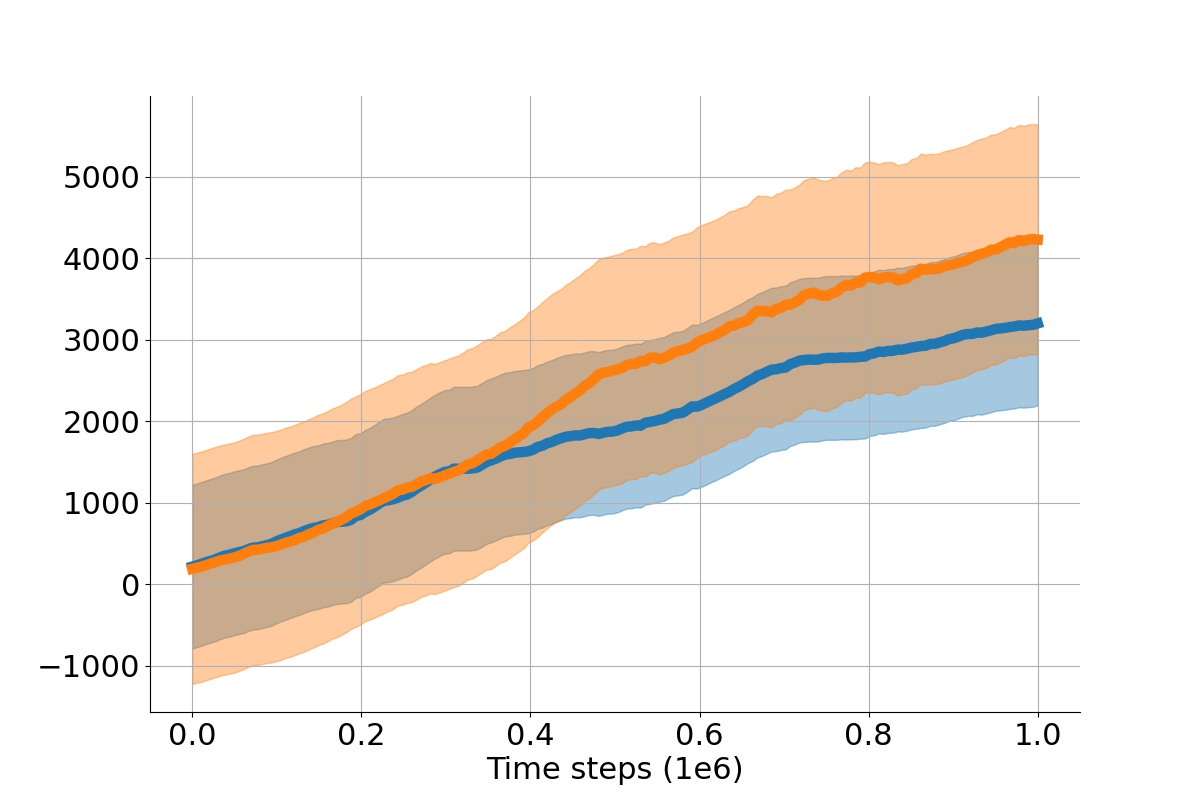}
	}
	\caption{Learning curves for the set of OpenAI Gym continuous control tasks. The shaded region represents half a standard deviation of the average evaluation over ten trials. Curves are smoothed uniformly with a sliding window of 20 for visual clarity.}
	\label{evaluation_results}
\end{figure*}

\subsection{Implementation Details}
For the implementation of TD3 \cite{td3}, we utilize the open-source implementation from the author's GitHub (\url{https://github.com/sfujim/TD3}). The TD3 \cite{td3} implementation that we use is the most recent and fine-tuned version of the algorithm updated by the author as of April 2021. Our modification is built on top of the TD3 \cite{td3} implementation such that the number of critics and target Q-value computation are replaced.

\subsection{Experimental Setup}
We perform true and estimated Q-value comparisons for our approach and TD3 \cite{td3} over 6 OpenAI Gym \cite{gym} continuous control tasks are presented in Fig. \ref{td3_tcd3_q_estimation}. Each task is run for 1 million time steps, and curves are derived through the same procedure as explained in section \ref{section:underestimation_problem}.

Our evaluations on each task are performed by rerunning both algorithms over 1 million time steps and evaluating the performance of the agent every 5000 time steps without exploration noise. Each evaluation report is the average of 10 episode rewards on the distinct evaluation environment. The results are reported over 10 random seeds of the Gym \cite{gym} simulator, network initialization, and code dependencies. 

\subsection{Discussion}
TCD3 obtains more accurate Q-value estimations than TD3 \cite{td3} on all environments. We observe two cases from our empirical results. First, our approach accurately estimates the Q-values by a negligible margin. Second, albeit the TCD3 greatly reduces the estimation bias, there still exists an underestimation error which is slightly bigger than half of the underestimation in TD3 \cite{td3}. These findings support our claim that an increasing variance of the reward signals encountered by the agent increases the underestimation, for example, in the Ant environment. Nonetheless, the expected value of the bias can be further reduced, as shown in this paper. 

The evaluation results on the same set of tasks are depicted by Fig. \ref{evaluation_results}. Our algorithm either outperforms or matches the TD3 algorithm \cite{td3} in terms of both learning speed and final performance. We observe that the estimation error prevents the agent from reaching higher possible reward potentials and stable returns. On top of the improvement that the minimum of two approximate critics can eliminate the overestimation bias, our approach obtains higher and smoother evaluation returns by reducing underestimation to a negligible margin and keeping the overestimation eliminated.

These simulation results also demonstrate that the underestimation induced by the minimum of two critics can cause ``good" state-action values to be assigned low values, resulting in slower convergence and suboptimal action choices to be selected more frequently \cite{off_policy}. For example, from Fig. \ref{td3_tcd3_q_estimation} and \ref{evaluation_results}, we observe that for BipedalWalker and Humanoid environments, elimination of estimation bias yields faster convergence to the optimal evaluation returns. Overall, by significantly reducing this underestimation phenomenon for deterministic continuous control actor-critic methods, we show in this paper that agents can attain higher evaluation returns in fewer time steps with no estimation bias.

\section{Conclusion}
Accumulated overestimation bias induced by the function approximation in deep reinforcement learning has been identified as a substantial issue. On the contrary, in deterministic actor-critic settings, techniques to overcome function approximation error build-up may lead to underestimation bias which has been a problematic drawback. In this work, we first show that encountering different reward signals increases the underestimation bias. Then, we develop a novel variant of deep Q-learning that significantly reduces the underestimation bias to a negligible level. Taken our claim and empirical results together, this improvement defines our efficient, parameter-free update rule, Triplet Critic Update, which dramatically improves both the learning speed and performance of TD3 in several challenging continuous control tasks. Our algorithm, Triplet Critic Delayed Deep Deterministic Policy Gradient (TCD3), outperforms state-of-the-art by obtaining more accurate Q-value estimates. Our modification is orthogonal and can easily be adapted to any deterministic actor-critic method that employs temporal difference learning.

\ifCLASSOPTIONcaptionsoff
  \newpage
\fi

\bibliographystyle{IEEEtran}
\bibliography{main}

\end{document}